\documentclass[10pt,twocolumn,letterpaper]{article}

\usepackage[pagenumbers]{cvpr} 

\usepackage{multirow}
\usepackage[misc]{ifsym}

\newcommand\blfootnote[1]{%
  \begingroup
  \renewcommand\thefootnote{}\footnote{#1}%
  \addtocounter{footnote}{-1}%
  \endgroup
}

\definecolor{cvprblue}{rgb}{0.21,0.49,0.74}
\usepackage[pagebackref,breaklinks,colorlinks,allcolors=cvprblue]{hyperref}

\def\paperID{2910} 
\def\confName{CVPR}
\def\confYear{2025}

\title{Multiple Object Tracking as ID Prediction}

\author{Ruopeng Gao\textsuperscript{1} \qquad Ji Qi\textsuperscript{2} \qquad Limin Wang\textsuperscript{1,3, {\footnotesize \Letter}} \\
    $^1$State Key Laboratory for Novel Software Technology, Nanjing University \\ $^2$China Mobile (Suzhou) Software Technology Co., Ltd. \qquad $^3$Shanghai AI Lab \\
    {\tt\small ruopenggao@gmail.com, lmwang@nju.edu.cn}
}

\begin{document}
\maketitle
\begin{abstract}
Multi-Object Tracking (MOT) has been a long-standing challenge in video understanding. 
A natural and intuitive approach is to split this task into two parts: object detection and association.
Most mainstream methods employ meticulously crafted heuristic techniques to maintain trajectory information and compute cost matrices for object matching. Although these methods can achieve notable tracking performance, they often require a series of elaborate handcrafted modifications while facing complicated scenarios.
We believe that manually assumed priors limit the method's adaptability and flexibility in learning optimal tracking capabilities from domain-specific data.
Therefore, we introduce a new perspective that treats \textbf{M}ultiple \textbf{O}bject \textbf{T}racking as an in-context \textbf{I}D \textbf{P}rediction task, transforming the aforementioned object association into an end-to-end trainable task.
Based on this, we propose a simple yet effective method termed \textbf{MOTIP}. 
Given a set of trajectories carried with ID information, MOTIP directly decodes the ID labels for current detections to accomplish the association process.
Without using tailored or sophisticated architectures, our method achieves state-of-the-art results across multiple benchmarks by solely leveraging object-level features as tracking cues. The simplicity and impressive results of MOTIP leave substantial room for future advancements, thereby making it a promising baseline for subsequent research. 
Our code and checkpoints are released at \href{https://github.com/MCG-NJU/MOTIP}{https://github.com/MCG-NJU/MOTIP}.
\end{abstract}  

\blfootnote{\Letter~: Corresponding author.}

\section{Introduction}
\label{sec:intro}

\begin{figure}[t]
    \centering
    \includegraphics[width=0.95\linewidth]{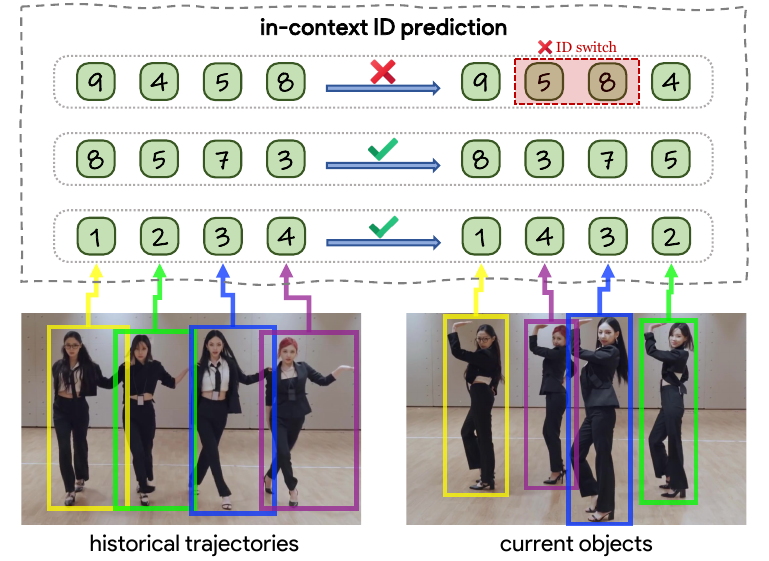}
    \caption{Diagram of the in-context ID prediction process. Different colored bounding boxes represent targets corresponding to different trajectories. We provide two valid ID prediction results, shown in the two lines below. This indicates that each trajectory only needs to predict the corresponding label based on the historical ID information, rather than being assigned a fixed label.}
\label{fig:in-context-prediction}
\end{figure}

The objective of multiple object tracking (MOT) is to accurately locate all objects of interest within a video stream while consistently maintaining their respective identities throughout the sequence. 
As an essential problem in computer vision, it is crucial for many downstream tasks, such as action recognition~\cite{Unified-MOT-Action} and trajectory prediction~\cite{Joint-MOT-Trajectory-Prediction}.
In practical applications, it has also played a substantial role in various fields, including autonomous driving~\cite{BDD100K}, sports event analysis~\cite{SportsMOT, TeamTrack, SportsHHI}, animal behavior research~\cite{NetTrack}, and so on.
Consequently, the challenges and advancements in multiple object tracking (MOT) have long garnered attention from the community. 

In the early stages of research within multi-object tracking area, the application scenarios and benchmarks were largely concentrated on pedestrian tracking~\cite{MOT15, MOT16}. In this scenario, the characteristics of pedestrians are primarily linear motion and distinguishable appearance. Therefore, at that time, the methods~\cite{SORT, ByteTrack} predominantly relied on the Kalman filter~\cite{KF} to model trajectories and predict their locations in the current frame, subsequently employing manually-designed algorithms for target matching.
Subsequent research~\cite{Deep-SORT, FairMOT, JDE} introduced additional re-identification modules to compute the similarity between trajectories and current objects, aiding in resolving long-term occlusions that are challenging for linear motion estimations. 
Despite achieving notable success in pedestrian tracking, these methods have struggled to keep up with the emergence of increasingly complex tracking scenarios~\cite{DanceTrack, SportsMOT, NetTrack}. 
In these scenarios, irregular movements and similar appearances deviate from heuristic priors, reducing the effectiveness of fixed matching rules then weakening tracking performance.
Although some of the latest heuristic algorithms~\cite{Hybrid-SORT, DeconfuseTrack, Deep-OC-SORT} can gradually adapt to these cases, the compromise is that each improvement requires substantial human analysis as well as meticulous tuning of rules and hyperparameters.

In recent years, some scholars have proposed end-to-end trainable MOT methods~\cite{MeMOT, CenterTrack, TransCenter, MOTR, TrackFormer} that directly learn tracking capabilities from the given training data to pursue the optimal solution. 
Among these, the methods~\cite{TransTrack, TrackFormer, MOTR, MeMOTR, SambaMOTR} based on extending DETR~\cite{DETR, DeformableDETR, DAB-DETR} to MOT have garnered significant attention and research. They propagate track queries across video frames to represent different trajectories. 
Despite achieving impressive results on multiple benchmarks, particularly on some highly challenging ones, these methods still leave some concerning issues. 
The most notable is that simultaneously using different types of queries for detection and tracking can cause conflicts within the unified decoding process~\cite{CO-MOT, MOTRv3}, thus impairing either detection or tracking performance. 
Some studies~\cite{MOTRv2} have found that using an additional independent detector~\cite{YOLOX} to decouple multi-object tracking can effectively mitigate this issue. Other studies~\cite{CO-MOT, MOTRv3} have also shown that handling detection and tracking within the same module can lead to conflicts in the allocation of supervision signals.
Reflecting on the above, a natural question arises: 
\textit{Can we maintain the decoupling nature of the multi-object tracking problem while discarding heuristic algorithms in favor of an end-to-end pipeline to fully unleash the model's potential?}

Since there are already many mature end-to-end frameworks for object detection~\cite{DETR, YOLOX, DeformableDETR}, we primarily focus on the formulation of object association. 
Intuitively, it resembles a classification problem, as different trajectories are annotated with distinct labels. 
However, considering the generalization to unseen trajectories during inference, \ie, new ID labels, classification prediction cannot be directly applied to object association. 
This is why, despite some ReID-based methods~\cite{FairMOT} using label classification for supervision, cosine similarity is employed during inference to calculate the affinity matrix, thereby determining the association results.
We reflect on this generalization dilemma, which arises because the labels of trajectories differ from traditional classification tasks~\cite{ImageNet}. Although a unique number annotates each trajectory as its ID, this does not imply that it can only be represented as this label. On the contrary, it is considered acceptable as long as a trajectory is predicted with the same ID label at all time steps.
Therefore, we consider treating the object association problem as an in-context ID prediction problem, as illustrated in~\cref{fig:in-context-prediction}.
Specifically, for a target in the current frame, we only need to predict its ID label based on the ID information carried by the corresponding historical trajectory, rather than predicting a globally fixed label as in traditional classification tasks.  
This ensures the generalization ability while facing unseen identities during inference.
In this way, the target association is formulated into a novel framework, maintaining consistency and end-to-end in both training and inference.

Based on the perspective above, we propose our method, \textbf{MOTIP}, by treating \textbf{M}ultiple \textbf{O}bject \textbf{T}racking as an \textbf{I}D \textbf{P}rediction problem.
Specifically, we opted for Deformable DETR~\cite{DeformableDETR} as our detector because it can directly provide object-level embeddings while detecting targets, without the need to consider various feature extraction techniques such as RoI, hierarchical structures, or feature pooling.
To represent the identity information for each trajectory, we store a set of learnable ID embeddings, which are attached to specific trajectory tokens as needed. 
As for the crucial ID prediction module, we simply use a standard transformer decoder~\cite{Attention}, composed of multiple layers of alternating self-attention and cross-attention.
Despite our minimalist and straightforward design, without employing tailored and sophisticated network structures, it demonstrates impressive state-of-the-art tracking performance across multiple benchmarks.
Therefore, we believe that framing multi-object tracking as an ID prediction problem still holds significant untapped potential, which can be further explored in future research.

\section{Related Work}
\label{sec:related}

\vspace{2pt}
\noindent \textbf{Tracking-by-Detection} 
is the most widely used paradigm for multi-object tracking in the community. These methods~\cite{ByteTrack, SORT, AgriSORT} employ post-processing strategies to associate detection results with historical trajectories, thereby achieving online multiple object tracking frame by frame. 
Most of them~\cite{SORT, ByteTrack} rely on Kalman filter~\cite{KF} to handle linear pedestrian motion~\cite{MOT15, MOT16, DIVOTrack} and leverage ReID features~\cite{Deep-SORT, DBLP:conf/icmcs/ChenAZS18, DBLP:journals/mta/MahmoudiAR19, FairMOT, JDE, Deep-OC-SORT, HATReID-MOT} to incorporate object appearance cues.
In recent years, many methods~\cite{OC-SORT, C-BIoU, BoT-SORT, Hybrid-SORT, TrackFlow, DeconfuseTrack, SparseTrack, LongTailMOT, GHOST, QuoVadis, UMT, OccTrack, MOTSync, QuoVadis} have adopted more complicated modeling and matching approaches or introduced additional multimodal information to mitigate the limitations of manual algorithms in complex scenarios~\cite{DanceTrack, SportsMOT}.
Our proposed MOTIP also structurally decouples detection and association, but it relies on learnable models rather than heuristic algorithms.
While some modern approaches~\cite{MotionTrack, MambaTrack+, MambaTrack, DiffMOT, DiffusionTrack} also utilize learnable modules to capture motion patterns, they still depend on handcrafted decisions to accomplish object association.
In contrast, our method incorporates the decision-making process into the end-to-end pipeline, which significantly highlights the uniqueness of our approach.

\vspace{2pt}
\noindent \textbf{Tracking-by-Propagation} 
is a recently popular end-to-end multiple object tracking paradigm. 
Inspired by query-based detection models~\cite{DETR, DAB-DETR, DeformableDETR}, they~\cite{MOTR, TrackFormer} extended the detect queries to the tracking task, using track queries to represent tracked targets and propagate them through the video sequence. There are also some methods~\cite{MQT, MeMOT} that employ tailored query-based model variants. 
Due to the flexibility of end-to-end trainable models, they are more adaptable to tracking in complex scenarios~\cite{DanceTrack}.
Subsequent works~\cite{MeMOT, MeMOTR, SambaMOTR} have focused on long-term modeling, further enhancing the training performance.
Nevertheless, some studies~\cite{CO-MOT, MOTRv3, MOTRv2} have pointed out irreconcilable conflicts in the process of joint detection and tracking and attempted to alleviate this issue.
Although our MOTIP also utilizes a query-based detection model~\cite{DeformableDETR}, we do not fall into this paradigm because our object detection and association are performed sequentially in two separate modules, rather than simultaneously. 

\section{Method}
\label{sec:method}

\begin{figure*}[t]
    \centering
    \includegraphics[width=0.95\linewidth]{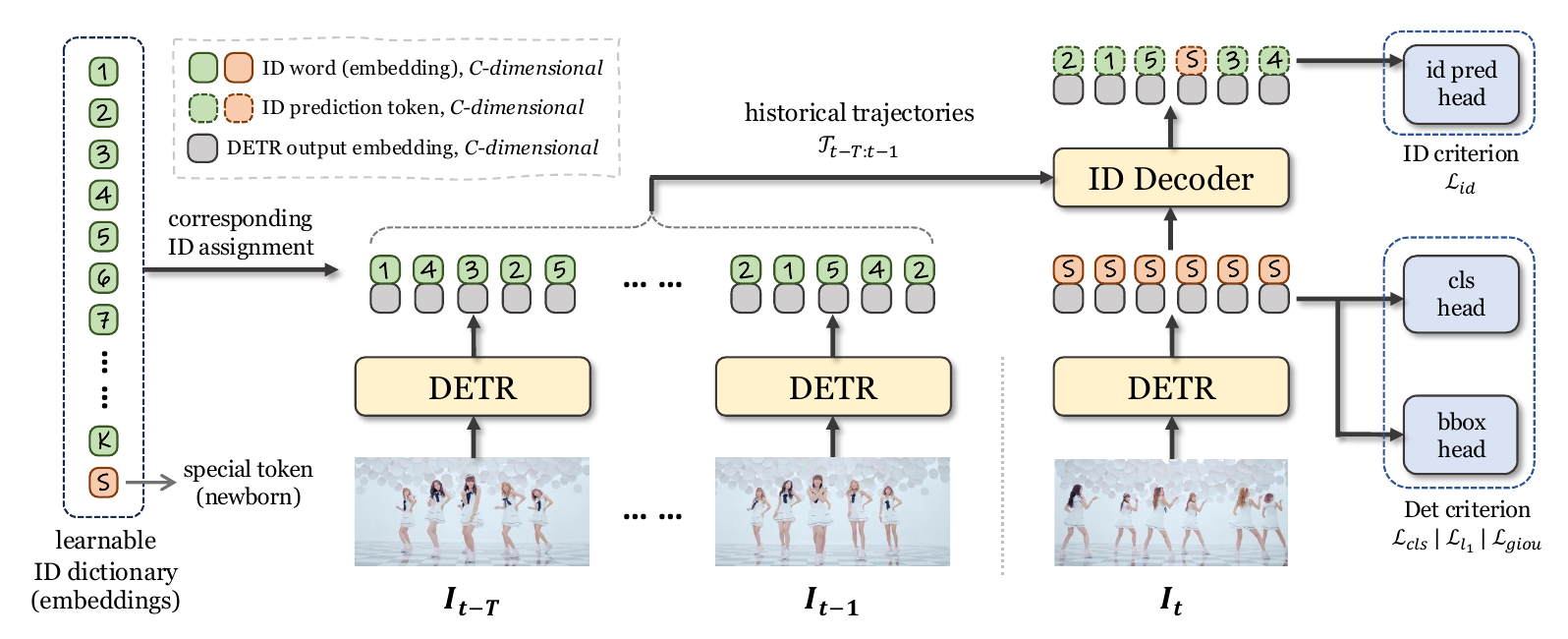}
    \caption{\textbf{Overview of MOTIP.} There are three primary components: a DETR detector detects objects, a learnable ID dictionary represents different identities, and an ID Decoder predicts the ID labels of current objects, as we detailed in~\cref{sec:motip-architecture}. We combine object features with their corresponding ID embeddings to form the historical trajectories $\mathcal{T}_{t-T:t-1}$. Subsequently, the ID tokens are regarded as identity prompts, and the ID Decoder performs in-context ID prediction based on them, as discussed in~\cref{sec:in-context-ID-prediction} and~\cref{sec:motip-architecture}.}
\label{fig:overview}
\end{figure*}

In this section, we detail our proposed method, MOTIP, which treats multiple object tracking as an in-context ID prediction task.
Firstly, in~\cref{sec:in-context-ID-prediction}, we introduce a novel perspective on how to formulate object association in MOT as an ID prediction problem. 
Subsequently, in~\cref{sec:motip-architecture}, we provide a detailed explanation of each component within MOTIP.
Finally, in~\cref{sec:training} and~\cref{sec:inference}, we further illustrate the processes of training and inference.

\subsection{In-context ID Prediction}
\label{sec:in-context-ID-prediction}

In multiple object tracking data, different trajectories are annotated with distinct ID labels. Therefore, some works~\cite{FairMOT} adopt classification loss to directly supervise the model in distinguishing different identities. 
However, during inference, the model will encounter unseen trajectories, which means it needs to predict out-of-distributed labels, leading to generalization issues. As a result, additional post-processing steps must be employed to complete the inference, such as using cosine similarity to determine object-matching results. 

Upon deep reflection, we believe this is due to the difference between ID labels in MOT and traditional classification tasks~\cite{ImageNet}.
In MOT, the labels of the trajectories are actually used to indicate a certain consistency rather than specific semantic information.
In other words, for a trajectory, as long as the ID label remains consistent across each frame, it is acceptable and does not need a specific label.
For example, in~\cref{fig:in-context-prediction}, the ID labels of these four objects are marked as \textit{1 2 3 4} in the ground truth file. However, we can also use \textit{8 5 7 3} to represent them. As long as the labels remain consistent in subsequent frames (as shown on the right), it will be considered a correct result.

Based on the above analysis, MOT can be regarded as a special label prediction problem where target labels are determined by historical trajectory identity information. 
Let $\mathcal{T}_{t-1}$ represent the historical trajectories, where $\mathcal{T}_{t-1} = \{\mathcal{T}_{t-1}^1, \mathcal{T}_{t-1}^2, \cdots, \mathcal{T}_{t-1}^M\}$ with each $\mathcal{T}_{t-1}^m$ representing a trajectory with a consistent identity. 
Simultaneously, we randomly assign an ID label $k_m$ to each trajectory $\mathcal{T}_{t-1}^m$ ensuring $1 \leq k_m \leq K$.
When a new frame $I_t$ is input, for any detected object $o_t$, if it belongs to the $m$-th trajectory $\mathcal{T}^m$, its correct ID label prediction result should be $k_m$. 
Since this prediction objective is based on the identity information $k$ attached to the historical trajectories, we refer to it as \textit{in-context ID prediction}, where the assigned label $k$ serves as an in-context prompt.
In~\cref{fig:in-context-prediction}, we have provided some examples.
Due to this formulation, the ID prediction results for any unseen trajectories will remain within the distribution of the training procedure, \ie, $1 \leq k \leq K$, thereby addressing the generalization dilemma.

\subsection{MOTIP Architecture}
\label{sec:motip-architecture}

The overall architecture of MOTIP is surprisingly simple, as shown in~\cref{fig:overview}. It contains three main components, which we will detail below: 
\textit{a DETR~\cite{DeformableDETR} detector} to detect objects and extract their object-level features, \textit{a learnable ID dictionary} to represent different in-context identity information, and \textit{an ID Decoder} to predict ID labels based on historical trajectories. 

\vspace{2pt}
\noindent \textbf{DETR Detector.}
We use Deformable DETR~\cite{DeformableDETR}, an end-to-end object detection model, as our detector. 
Starting from an input image $I_t$, the CNN~\cite{ResNet} backbone and transformer encoder extract and enhance the image features. 
Subsequently, the transformer decoder generates the output embeddings from learnable detect queries. They are decoded into bounding boxes and classification confidence by the \textit{bbox} and \textit{cls} head, as illustrated in~\cref{fig:overview}.
This approach further simplifies our method, as we can directly use the decoded output embedding as the target feature $f_t^n$, eliminating the need for complicated feature extraction techniques such as RoI, hierarchical methods, \etc.

\vspace{2pt}
\noindent \textbf{ID Dictionary.}
As discussed in~\cref{sec:in-context-ID-prediction}, for the model's generalization on unseen trajectories, we require additional signifiers to represent the identity information of trajectories, which are used as in-context prompts.
Since identity is discrete information, a na\"ive approach would be to use one-hot encoding. However, we believe this is not a good idea. Firstly, one-hot encoding is not conducive to neural network training. Secondly, this encoding scheme limits the number of ID labels to the vector dimensions that the model can handle, which is unfavorable for subsequent expansion and generalization.
Therefore, we create an ID dictionary $\mathcal{I}$ that consists of $K+1$ learnable words to represent different identities, as follows:

\begin{equation}
    \mathcal{I} = \{ i^1, i^2, \cdots, i^K, i^{\textit{spec}} \},
\label{eq:ID-dict}
\end{equation}
where each word $i^k$ is a learnable $C$-dimensional embedding. In detail, the first $K$ tokens $\{ i^1, i^2, \cdots, i^K \}$ are regular tokens that represent specific identities, while the last word $i^{\textit{spec}}$ is a \textit{special} token that stands for newborn objects.

\vspace{2pt}
\noindent \textbf{Tracklet Formation.}
In MOTIP, we only use object-level features as tracking cues. 
Therefore, for the $m$-th trajectory, we retain all target features from the past $T$ frames, denoted as $\mathcal{F}_{t-T:t-1} = \{ f_{t-T}^m, \cdots, f_{t-1}^m \}$, and randomly assign a unique ID label $k_m$. 
Then, we fuse the corresponding ID words $i^{k_m}$ with the target features $f_t^m$, so that the tracklets carry both tracking cues and in-context identity prompts needed for ID prediction, as discussed in~\cref{sec:in-context-ID-prediction}.
Here, we simply use concatenation to achieve this, as shown below:

\begin{equation}
    \tau_t^{m,k_m} = \textit{concat}(f_t^m, i^{k_m}).
\label{eq:tracklet}
\end{equation}
Here, $f_t^m$ is the C-dimensional output embedding from DETR, and $i^{k_m}$ is a $C$-dimensional token obtained from the dictionary~\cref{eq:ID-dict}. Ultimately, this results in a $2C$-dimensional tracklet representation $\tau_t^{m,k_m}$.
According to this, we denote all historical trajectories as $\mathcal{T}_{t-T:t-1} = \{ \cdots, \mathcal{T}_{t-T:t-1}^m, \cdots \}$, where $\mathcal{T}_{t-T:t-1}^m = \{ \tau_{t-T}^{m,k_m}, \cdots, \tau_{t-1}^{m,k_m} \}$.
For the sake of consistency, we apply the same construction form from~\cref{eq:tracklet} to the targets in the current frame. However, since there is no trajectory identity yet, we use the special token $i^{\textit{spec}}$ instead of $i^{k_m}$, denoted as $\tau_t^{n} = \textit{concat}(f_t^n, i^{\textit{spec}})$.

\vspace{2pt}
\noindent \textbf{ID Decoder.}
Due to the variable length and number of historical trajectories, we use a standard transformer decoder structure~\cite{Attention} as our ID Decoder to handle the variable-length inputs.
This component uses all historical tracklets $\tau^{m,k_m}$ as \textit{Key} and \textit{Value} to decode all active detection tracklets $\tau_t^n$ in the current frame, as illustrated in~\cref{fig:overview}.
We use a simple linear classification head to predict the ID label for the decoded output embeddings. 
If a detection $\tau_t^{n}$ belongs to the $m$-th trajectory, the classification head should predict it as $k_m$, since $i^{k_m}$ corresponds to the in-context ID information for that trajectory. 
This way, the entire object association process can be formulated as a classification task, allowing for direct supervision using cross-entropy loss.

\subsection{Training}
\label{sec:training}

\noindent \textbf{Loss Function.}
As previously discussed, we transform the object association in MOT into an end-to-end learnable $K+1$ classification problem through in-context ID prediction. 
Consequently, we can use the standard cross-entropy loss function as supervision, denoted as $\mathcal{L}_{\textit{id}}$.
Since DETR~\cite{DETR, DeformableDETR} can also be trained end-to-end, the entire MOTIP model can utilize a unified loss function $\mathcal{L}$ for supervision:

\begin{equation}
    \mathcal{L} = \lambda_{\textit{cls}} \mathcal{L}_{\textit{cls}} + \lambda_{\textit{L1}} \mathcal{L}_{\textit{L1}} + \lambda_{\textit{giou}} \mathcal{L}_{\textit{giou}} + \lambda_{\textit{id}} \mathcal{L}_{\textit{id}},
\label{eq:loss}
\end{equation}
where $\mathcal{L}_{\textit{cls}}$ is the focal loss~\cite{Focal-Loss}. $\mathcal{L}_{\textit{L1}}$ and $\mathcal{L}_{\textit{giou}}$ denote the L1 loss and the generalized IoU loss~\cite{G-IoU}, respectively. 
$\lambda_{\textit{cls}}$, $\lambda_{\textit{L1}}$ and $\lambda_{\textit{giou}}$ are their corresponding weight coefficients, and $\lambda_{\textit{id}}$ is the weight coefficient of ID loss $\mathcal{L}_{\textit{id}}$.

\begin{figure}[t]
    \centering
    \includegraphics[width=0.9\linewidth]{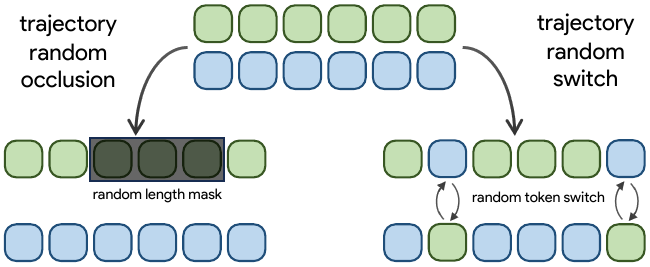}
    \caption{Illustration of trajectory augmentation: \textit{trajectory random occlusion} (left) and \textit{trajectory random switch} (right). Two different colors represent two distinct trajectories.}
\label{fig:trajectory-augmentation}
\end{figure}

\vspace{2pt}
\noindent \textbf{Trajectory Augmentation.}
In multiple object tracking, we often face numerous challenges, such as target occlusion, blurriness, and high similarity between targets. 
This can potentially lead to partial errors in ID assignment during online inference, which in turn can reduce the reliability of historical trajectories in subsequent processes. 
However, such errors do not occur during training because we use ground truth for supervision and trajectory construction. 
We argue that the oversimplification during training may prevent the model from acquiring sufficiently generalized and robust tracking capabilities. 
To mitigate this issue, we propose two trajectory augmentation techniques to be used during the training phase. 
Firstly, considering that occlusion is a challenging problem faced by MOT, we randomly drop tokens from each trajectory with a probability of $\lambda_{\text{occ}}$, as shown on the left of~\cref{fig:trajectory-augmentation}.
Secondly, considering the potential ID assignment errors during inference, we randomly swap the ID tokens of two trajectories within the same frame with a probability of $\lambda_{\text{sw}}$ to simulate the model assigning incorrect IDs to similar targets, as shown on the right of~\cref{fig:trajectory-augmentation}.

\subsection{Inference}
\label{sec:inference}

As discussed in~\cref{sec:in-context-ID-prediction} and~\cref{sec:motip-architecture}, during the inference stage, we can randomly assign an ID label $k_m$ to each trajectory $\mathcal{T}^m$ and use the corresponding ID embedding $i^{k_m}$, as long as the labels are unique across different trajectories. \Ie, for any two trajectories $\mathcal{T}^m$ and $\mathcal{T}^n$, $k_m \neq k_n$.
In the implementation, we sequentially assign ID labels from $1$ to $K$ to represent the trajectories.
For some longer video sequences, as trajectories expire and new ones appear, there may be more than $K$ trajectories.
To address this, we recycle the ID labels of the concluded trajectories for reuse.

In practice, for all output embeddings decoded by DETR, we first filter them using a detection confidence threshold $\lambda_\text{det}$. After that, all active detections are fed into the ID Decoder to predict the probability of each ID label. 
Similar to traditional classification tasks~\cite{ImageNet, ResNet}, for each object, we select the ID with the highest probability ($> \lambda_\text{id}$) as the final result.
Subsequently, if an object is not assigned a valid ID and its detection confidence is greater than $\lambda_\text{new}$, it will be marked as a newborn target and assigned a new identity.
This makes our inference process very simple and straightforward. 
Notably, due to the restriction against duplicate ID predictions in the evaluation process of MOT tasks, when two targets in the same frame are predicted to the same ID label, only the one with the highest confidence score is retained.
The pseudocode and additional details are provided in \cref{sec:supp-inference}.

From past experience, more complex or advanced ID assignment strategies, such as Hungarian algorithm or multi-stage matching~\cite{ByteTrack}, might offer some improvements. 
However, to validate the robustness and generalization of the model itself, we do not focus on these approaches.

\section{Experiments}
\label{sec:exps}

\subsection{Datasets and Metrics}
\label{sec:datasets-and-metrics}

\noindent \textbf{Datasets.}
To evaluate MOTIP, we select a variety of challenging benchmarks. 
DanceTrack~\cite{DanceTrack} is a multi-person tracking dataset composed of $100$ videos of various types of group dances. 
SportsMOT~\cite{SportsMOT} is a dataset focused on athlete tracking, composed of $240$ sports broadcast videos.
BFT~\cite{NetTrack} is a high-maneuverability target tracking dataset that includes $22$ bird species from around the world, consisting of $106$ video clips.
These benchmarks feature numerous serious challenges commonly face in multi-object tracking, such as frequent occlusions, irregular movements, high-speed motion, and similar appearance. 
This will help us fully verify the robustness and generalization ability of MOTIP in different scenarios.

\vspace{2pt}
\noindent \textbf{Metrics.}
We mainly use the Higher Order Tracking Accuracy (HOTA)~\cite{HOTA} to evaluate our method since it provides a balanced way to measure both object detection accuracy (DetA) and association accuracy (AssA).
We also list the MOTA~\cite{MOTA} and IDF1~\cite{IDF1} metrics in our experiments.

\subsection{Implementation Details}
\label{sec:implementation-details}

\noindent \textbf{Network.}
In practice, we select Deformable DETR~\cite{DeformableDETR} with a ResNet-50~\cite{ResNet} backbone as our default DETR detector because it is a versatile option for downstream tasks~\cite{MOTR, ConstrastiveDETR-MOT}. Similar to previous work~\cite{MeMOTR, CO-MOT, MOTRv3}, we also utilize the COCO~\cite{COCO} pre-trained weights as initialization.
We apply relative position encoding in the ID Decoder because tracking focuses more on relative temporal relationships rather than absolute timestamps.
To minimize unnecessary additional modules, the hidden dimension throughout the entire model is kept consistent with Deformable DETR, which is $C = 256$. 
Since the ID dictionary can be reused, it is only necessary to ensure that $K$ is not less than the maximum number of targets per frame. Here, we set $K$ to $50$ for simplicity.

\begin{table}[t]
  \centering
  \setlength{\tabcolsep}{2pt}{
    \begin{tabular}{l|ccccc}
      \toprule[2pt]
      Methods & HOTA & DetA & AssA & MOTA & IDF1 \\
      \midrule[1pt]
      \textit{w/o extra data:} \\
      FairMOT~\cite{FairMOT} & 39.7 & 66.7 & 23.8 & 82.2 & 40.8 \\
      CenterTrack~\cite{CenterTrack} & 41.8 & 78.1 & 22.6 & 86.8 & 35.7 \\
      TraDeS~\cite{TraDeS} & 43.3 & 74.5 & 25.4 & 86.2 & 41.2 \\
      TransTrack~\cite{TransTrack} & 45.5 & 75.9 & 27.5 & 88.4 & 45.2 \\
      ByteTrack~\cite{ByteTrack} & 47.7 & 71.0 & 32.1 & 89.6 & 53.9 \\
      GTR~\cite{GTR} & 48.0 & 72.5 & 31.9 & 84.7 & 50.3 \\
      QDTrack~\cite{QDTrack} & 54.2 & 80.1 & 36.8 & 87.7 & 50.4 \\
      MOTR~\cite{MOTR} & 54.2 & 73.5 & 40.2 & 79.7 & 51.5 \\
      OC-SORT~\cite{OC-SORT} & 55.1 & 80.3 & 38.3 & 92.0 & 54.6 \\
      StrongSORT~\cite{StrongSORT} & 55.6 & 80.7 & 38.6 & 91.1 & 55.2 \\
      C-BIoU~\cite{C-BIoU}& 60.6 & 81.3 & 45.4 & 91.6 & 61.6 \\
      Hybrid-SORT~\cite{Hybrid-SORT} & 62.2 & / & / & 91.6 & 63.0 \\
      DiffMOT~\cite{DiffMOT} & 62.3 & \bf 82.5 & 47.2 & \bf 92.8 & 63.0 \\
      MeMOTR~\cite{MeMOTR} & 63.4 & 77.0 & 52.3 & 85.4 & 65.5 \\
      CO-MOT~\cite{CO-MOT} & 65.3 & 80.1 & 53.5 & 89.3 & 66.5 \\
      MOTIP (ours) & \bf 69.6 & 80.4 & \bf 60.4 & 90.6 & \bf 74.7 \\
      \midrule[1pt]
      \textit{\textcolor{gray}{with extra data:}} \\
      \textcolor{gray}{MOTRv3~\cite{MOTRv3}} & \textcolor{gray}{68.3} & / & / & \textcolor{gray}{91.7} & \textcolor{gray}{70.1} \\
      \textcolor{gray}{CO-MOT~\cite{CO-MOT}} & \textcolor{gray}{69.4} & \textcolor{gray}{82.1} & \textcolor{gray}{58.9} & \textcolor{gray}{91.2} & \textcolor{gray}{71.9} \\
      \textcolor{gray}{MOTRv2~\cite{MOTRv2}} & \textcolor{gray}{69.9} & \bf \textcolor{gray}{83.0} & \textcolor{gray}{59.0} & \bf \textcolor{gray}{91.9} & \textcolor{gray}{71.7} \\
      \textcolor{gray}{MOTIP (ours)} & \bf \textcolor{gray}{72.0} & \textcolor{gray}{81.8} & \bf \textcolor{gray}{63.5} & \bf \textcolor{gray}{91.9} & \bf \textcolor{gray}{76.8} \\
      \bottomrule[2pt]
    \end{tabular}
  }
  \caption{Performace comparison with state-of-the-art methods on the DanceTrack~\cite{DanceTrack} test set. The best result is shown in \textbf{bold}.}
  \label{tab:sota-dancetrack}
\end{table}

\vspace{2pt}
\noindent \textbf{Training.}
As in the prior work~\cite{MeMOTR, MOTRv2}, we use several common data augmentation methods, such as random resize, crop, and color jitter. The shorter and longer side of the input image is resized to $800$ and $1440$, respectively. 

In each training iteration, we randomly sample $T+1$ frames of images with random intervals, perform ID prediction on the subsequent $T$ frames, and supervise them with~\cref{eq:loss}. 
To reduce computational costs, we only backpropagate the gradients for the DETR on four of these frames, while the remaining $T-3$ frames are processed in the no-gradient mode using \textit{torch.no\_grad()}.
By decoupling the detection and association problems and allowing the ID Decoder to use attention masks~\cite{Attention} to ensure future invisibility, our method can achieve high parallelism and be GPU-friendly. 
As a result, MOTIP can be efficiently trained using 8 NVIDIA RTX 4090 GPUs. For instance, training on DanceTrack~\cite{DanceTrack} takes less than one day.
More details and discussions about the training setups can be found in \cref{sec:supp-training}.

\vspace{2pt}
\noindent \textbf{Hyperparameters.}
In our experiments, the supervision weight coefficients $\lambda_{\textit{cls}}$, $\lambda_{\textit{L1}}$, $\lambda_{\textit{giou}}$ and $\lambda_{\textit{id}}$ are set to $2.0$, $5.0$, $2.0$ and $1.0$. 
The maximum temporal length $T$ is set to $29$, $59$, and $19$ for DanceTrack, SportsMOT, and BFT, respectively. 
The inference thresholds $\lambda_\text{det}$, $\lambda_\text{new}$, and $\lambda_\text{id}$ are set to $0.3$, $0.6$, $0.2$.
For the training augmentation parameters, we set $\lambda_\text{occ} = \lambda_\text{sw} = 0.5$.
Although fine-tuning some hyperparameters on different datasets may yield better results, for simplicity, we strive to maintain their consistency.

\begin{table}[t]
  \centering
  \setlength{\tabcolsep}{2pt}{
    \begin{tabular}{l|ccccc}
      \toprule[2pt]
      Methods & HOTA & DetA & AssA & MOTA & IDF1 \\
      \midrule[1pt]
      \textit{w/o extra data:} \\
      FairMOT~\cite{FairMOT} & 49.3 & 70.2 & 34.7 & 86.4 & 53.5\\
      QDTrack~\cite{QDTrack} & 60.4 & 77.5 & 47.2 & 90.1 & 62.3\\
      ByteTrack~\cite{ByteTrack} & 62.1 & 76.5 & 50.5 & 93.4 & 69.1 \\
      TrackFormer~\cite{TrackFormer} & 63.3 & 66.0 & 61.1 & 74.1 & 72.4 \\
      OC-SORT~\cite{OC-SORT} & 68.1 & \bf 84.8 & 54.8 & \bf 93.4 & 68.0 \\
      MeMOTR~\cite{MeMOTR} & 68.8 & 82.0 & 57.8 & 90.2 & 69.9 \\
      MOTIP (ours) & \bf 72.6 & 83.5 & \bf 63.2 & 92.4 & \bf 77.1 \\
      \midrule[1pt]
      \textit{\textcolor{gray}{with extra data:}} \\
      \textcolor{gray}{GTR~\cite{GTR}} & \textcolor{gray}{54.5} & \textcolor{gray}{64.8} & \textcolor{gray}{45.9} & \textcolor{gray}{67.9} & \textcolor{gray}{55.8} \\
      \textcolor{gray}{CenterTrack~\cite{CenterTrack}} & \textcolor{gray}{62.7} & \textcolor{gray}{82.1} & \textcolor{gray}{48.0} & \textcolor{gray}{90.8} & \textcolor{gray}{60.0} \\
      \textcolor{gray}{ByteTrack~\cite{ByteTrack}} & \textcolor{gray}{62.8} & \textcolor{gray}{77.1} & \textcolor{gray}{51.2} & \textcolor{gray}{94.1} & \textcolor{gray}{69.8} \\
      \textcolor{gray}{TransTrack~\cite{TransTrack}} & \textcolor{gray}{68.9} & \textcolor{gray}{82.7} & \textcolor{gray}{57.5} & \textcolor{gray}{92.6} & \textcolor{gray}{71.5} \\
      \textcolor{gray}{OC-SORT~\cite{OC-SORT}} & \textcolor{gray}{71.9} & \textcolor{gray}{86.4} & \textcolor{gray}{59.8} & \textcolor{gray}{94.5} & \textcolor{gray}{72.2} \\
      \textcolor{gray}{DiffMOT~\cite{DiffMOT}} & \textcolor{gray}{72.1} & \textcolor{gray}{86.0} & \textcolor{gray}{60.5} & \textcolor{gray}{94.5} & \textcolor{gray}{72.8} \\
      \bottomrule[2pt]
    \end{tabular}
  }
  \caption{Performace comparison with state-of-the-art methods on the SportsMOT~\cite{SportsMOT} test set. The best is shown in \textbf{bold}. The results of existing methods are from prior work~\cite{SportsMOT, MeMOTR, SambaMOTR}.}
  \label{tab:sota-sportsmot}
\end{table}

\subsection{Comparisons with State-of-the-art Methods}
\label{sec:sota}

We compare MOTIP with numerous previous methods on the DanceTrack~\cite{DanceTrack}, SportsMOT~\cite{SportsMOT}, and BFT~\cite{NetTrack} benchmarks, as shown in~\cref{tab:sota-dancetrack},~\cref{tab:sota-sportsmot}, and~\cref{tab:sota-bft}, respectively. 
For recent tracking-by-query methods~\cite{MOTR, MOTRv2} that also use DETR, studies~\cite{MeMOTR, MOTRv3, SambaMOTR} have shown that the choice of different DETR~\cite{DeformableDETR, DAB-DETR} and backbone~\cite{ResNet, ConvNeXT} networks can significantly impact performance. Therefore, we chose Deformable DETR~\cite{DeformableDETR} with a ResNet-50~\cite{ResNet} backbone as the competing platform to ensure a fair comparison. 
Some methods~\cite{MOTRv2, CO-MOT, MOTRv3, SportsMOT} use extra detection datasets to simulate video clips for joint training. We argue this approach is detrimental to the robustness of end-to-end, especially long-term modeling methods, as detailed and discussed in \cref{sec:supp-rethinking} and other research~\cite{MeMOTR, SambaMOTR}. Therefore, we primarily compare results without using additional datasets and still demonstrate superior performance.

\vspace{2pt}
\noindent \textbf{DanceTrack.}
The complex scenarios of frequent occlusions and irregular motion pose a severe challenge to heuristic algorithms~\cite{ByteTrack, FairMOT}. 
Methods such as Hybrid-SORT~\cite{Hybrid-SORT}, C-BIoU~\cite{C-BIoU}, and others~\cite{DiffMOT, StrongSORT}, despite utilizing a more powerful detector~\cite{YOLOX}, more intricate manual designs, and additional tracking cues to enhance performance, are still significantly outperformed by MOTIP. 
Compared to the strong competitor CO-MOT~\cite{CO-MOT}, which also uses Deformable DETR~\cite{DeformableDETR}, we achieve a new state-of-the-art result with a notable lead of $4.3$ HOTA and $6.9$ AssA, even surpassing some outstanding results~\cite{MOTRv3, CO-MOT} that using additional datasets for training (as shown in the lower of~\cref{tab:sota-dancetrack}). 
Such impressive performance demonstrates the considerable potential of our approach in extremely challenging scenarios.

\vspace{2pt}
\noindent \textbf{SportsMOT.}
Sports broadcasts involve frequent camera movements, accompanied by athletes' high-speed movements and repeated interactions.
OC-SORT~\cite{OC-SORT} effectively handles sudden stops and starts by explicitly modeling non-linear movements, resulting in a significant improvement over its predecessor~\cite{ByteTrack}.
In experiments, our proposed MOTIP significantly outperforms all previous methods by a considerable margin while also surpasses competitors~\cite{MeMOTR, TrackFormer} using the same detector~\cite{DeformableDETR}.
To avoid introducing additional engineering challenges and intricate remedies, as elaborated in \cref{sec:supp-rethinking}, we have not provided the results with extra training datasets like~\cite{DiffMOT, MotionTrack, Deep-EIoU}. 
However, our method, trained solely on the SportsMOT train set, still surpasses many joint training methods~\cite{OC-SORT, DiffMOT, TransTrack} especially on the association accuracy (AssA), as shown in the lower part of~\cref{tab:sota-sportsmot}, demonstrating our commendable performance and potential.

\begin{table}[t]
  \centering
  \setlength{\tabcolsep}{2pt}{
    \begin{tabular}{l|ccccc}
      \toprule[2pt]
      Methods & HOTA & DetA & AssA & MOTA & IDF1 \\
      \midrule[1pt]
      JDE~\cite{JDE} & 30.7 & 40.9 & 23.4 & 35.4 & 37.4 \\
      CSTrack~\cite{CSTrack} & 33.2 & 47.0 & 23.7 & 46.7 & 34.5 \\
      FairMOT~\cite{FairMOT} & 40.2 & 53.3 & 28.2 & 56.0 & 41.8 \\
      TransCenter~\cite{TransCenter} & 60.0 & 66.0 & 61.1 & 74.1 & 72.4 \\
      SORT~\cite{SORT} & 61.2 & 60.6 & 62.3 & 75.5 & 77.2 \\
      ByteTrack~\cite{ByteTrack} & 62.5 & 61.2 & 64.1 & \bf 77.2 & \bf 82.3 \\
      TrackFormer~\cite{TrackFormer} & 63.3 & 66.0 & 61.1 & 74.1 & 72.4 \\
      CenterTrack~\cite{CenterTrack} & 65.0 & 58.5 & 54.0 & 60.2 & 61.0 \\
      OC-SORT~\cite{OC-SORT} & 66.8 & 65.4 & 68.7 & 77.1 & 79.3 \\
      MOTIP (ours) & \bf 70.5 & \bf 69.6 & \bf 71.8 & 77.1 & 82.1 \\
      \bottomrule[2pt]
    \end{tabular}
  }
  \caption{Performace comparison with state-of-the-art methods on the BFT~\cite{NetTrack} test set. The best performance is shown in \textbf{bold}. The results of existing methods are derived from~\cite{NetTrack} and~\cite{SambaMOTR}.}
  \label{tab:sota-bft}
\end{table}

\vspace{2pt}
\noindent \textbf{BFT.}
Tracking birds differs in many ways from tracking humans~\cite{MOT16, MOT20, DanceTrack, SportsMOT, TeamTrack}. 
On the one hand, birds have highly dynamic movements due to their three-dimensional activity space, compared to ground targets. 
On the other hand, their appearance is often more similar due to the absence of artificial distinctions such as clothing.
Therefore, this presents a challenging new problem that is different from previous ones. 
Nevertheless, as shown in~\cref{tab:sota-bft}, our MOTIP has established a new state-of-the-art result with $70.5$ HOTA and $71.8$ AssA. 
This helps demonstrate the generalization ability of our method across different scenarios.

\subsection{Ablations}
\label{sec:ablation}

We conduct our ablation experiments on DanceTrack~\cite{DanceTrack} because it is challenging and offers a large-scale training set that better unlocks the model's potential. 
Unless otherwise stated, all trajectory augmentation techniques will not be used, \ie, $\lambda_\text{occ} = \lambda_\text{sw} = 0.0$. 
More details and analyses will be elaborated in \cref{sec:supp-ablation}.

\vspace{2pt}
\noindent \textbf{Hungarian Algorithm.}
The Hungarian algorithm is a commonly used approach for finding global optimal solutions~\cite{ByteTrack, OC-SORT}. 
However, by default, we do not use the Hungarian algorithm in our method, but rather opt for the more straightforward inference procedure described in~\cref{sec:inference}. 
Nonetheless, we explore its impact on our MOTIP.
As shown in the bottom half of~\cref{tab:self}, it does not provide considerable benefits to our method. 
We believe this is because our model inherently possesses the ability to find optimal solutions, which also indicates that our approach is far removed from traditional heuristic algorithms.

\vspace{2pt}
\noindent \textbf{Self-Attention in ID Decoder.}
Earlier, we mentioned that MOTIP can find global optimal solutions. We believe this can be attributed to the self-attention layers in the ID Decoder. 
We perform an ablation study on this design in~\cref{tab:self}. 
Not surprisingly, using only the decoder layers can still achieve acceptable tracking performance.
However, we argue that self-attention layers are crucial for better tracking. This is because they help the current objects exchange identity information during inference, thereby preventing confusion among similar targets. 
Therefore, the impact of the Hungarian algorithm is amplified, which is why you can observe a remarkable improvement.
When trajectory augmentation is introduced, the performance gap between the approach without self-attention layers and the final MOTIP further widens, underscoring the critical role of self-attention layers.

\begin{table}[t]
  \centering
  \setlength{\tabcolsep}{3.5pt}{
    \begin{tabular}{c|cc|ccccc}
    \toprule[2pt]
    \textit{self} & \textit{hung} & \textit{aug} & HOTA & DetA & AssA & MOTA & IDF1 \\ 
    \midrule[1pt]
    &&& 57.7 & \bf 76.3 & 43.9 & 85.5 & 56.1 \\
    & \checkmark & & 59.7 & 75.8 & 47.2 & \bf 85.6 & 59.9 \\
    & \checkmark & \checkmark & 60.8 & 75.5 & 49.3 & 83.7 & 62.5 \\
    & & \checkmark & 60.2 & 75.4 & 48.2 & 82.0 & 61.3 \\
    \midrule[1pt]
    \checkmark & & & 59.5 & 75.3 & 47.2 & 85.6 & 61.1 \\
    \checkmark & \checkmark & & 59.9 & 75.1 & 47.9 & 85.6 & 62.2 \\
    \checkmark & \checkmark & \checkmark & \bf 62.2 & 75.2 & \bf 51.8 & 85.4 & \bf 65.6 \\
    \rowcolor{gray!20} \checkmark & & \checkmark & \bf 62.2 & 75.3 & 51.5 & 85.2 & 64.8 \\
    \bottomrule[2pt]
    \end{tabular}
  }
  \caption{Evaluate the impact of different components and strategies. Let \textit{self}, \textit{hung}, and \textit{aug} symbolize the self-attention layer, Hungarian algorithm, and trajectory augmentation, respectively. The \colorbox{gray!20}{gray} background is the choice for our final experiment.}
  \label{tab:self}
\end{table}

\begin{table}[t]
  \centering
  \setlength{\tabcolsep}{5pt}{
    \begin{tabular}{c|c|c|ccc}
      \toprule[2pt]
      \# & \textit{Train} & \textit{Form} & HOTA & AssA & IDF1 \\
      \midrule[1pt]
       \# 1 & \multirow{3}{*}{\textit{Two-Stage}} & \textit{re-id} & 29.4 & 11.5 & 22.1 \\
       \# 2 & & \textit{contra} & 41.0 & 22.6 & 36.4 \\
       \# 3 & & \textit{id-pred} & 55.4 & 41.1 & 55.7 \\
      \midrule[1pt]
       \# 4 & \multirow{7}{*}{\textit{One-Stage}} & \textit{re-id} & 41.0 & 22.5 & 35.0 \\
       \# 5 & & \textit{re-id$^\ddag$} & 50.6 & 34.7 & 50.9 \\
       \# 6 & & \textit{contra} & 49.8 & 33.0 & 47.1 \\
       \# 7 & & \textit{contra$^\ddag$} & 52.6 & 37.3 & 54.0 \\
       \# 8 & & \textit{$\star$contra} & 51.2 & 35.0 & 48.1 \\
       \# 9 & & \textit{$\star$contra$^\ddag$} & 54.5 & 40.3 & 55.5 \\
       \rowcolor{gray!20} \# 10 & & \textit{id-pred} & \bf 59.5 & \bf 47.2 & \bf 61.1 \\
      \bottomrule[2pt]
    \end{tabular}
  }
  \caption{
  Comparison with common ReID pipelines. 
  As the tracking formulation (\textit{Form}), \textit{re-id} and \textit{contra} represent training the model using the formula from two well-known ReID methods,~\cite{FairMOT} and~\cite{ConstrastiveDETR-MOT}, respectively, and inference is based on cosine similarity.
  The \textit{id-pred} indicates our proposed MOTIP.
  $\ddag$ and $\star$ represent the use of the Hungarian algorithm and the trajectory enhancement module, respectively.
  }
  \label{tab:formula}
\end{table}

\vspace{2pt}
\noindent \textbf{Comparison with ReID Pipelines.}
Since our MOTIP also uses object-level features as tracking cues, it can easily be mistaken for a type of ReID method. 
In~\cref{tab:formula}, we compare two ReID learning pipelines derived from~\cite{FairMOT} and~\cite{ConstrastiveDETR-MOT}, identified as \textit{re-id} and \textit{contra}, respectively.
In the upper section (\#1 to \#3) of~\cref{tab:formula}, we perform experiments using the frozen, well-trained DETR weights. These results clearly illustrate that our method shows significant advantages over the other two formulations under the same object features.
In the remaining part of~\cref{tab:formula}, we jointly train all network parameters in a one-stage manner.
To eliminate the influence of introducing additional structures, we incorporate a trajectory enhancement module, identical to the structure of our ID Decoder, into some experiments, denoted as $\star$.
The experimental results demonstrate that, whether utilizing a trajectory enhancement module or an advanced assignment strategy (Hungarian algorithm, refer to $\ddag$), these methods still lag behind our MOTIP.
This can be attributed to MOTIP's ability to manage historical tracklets with greater flexibility, as visualized and discussed further in \cref{sec:supp-visualization-ID-decoder}.
In contrast, ReID methods~\cite{FairMOT, JDE, ConstrastiveDETR-MOT} employ heuristic algorithms to integrate historical information and perform similarity calculation independently, which limits the model's adaptability.
Furthermore, we emphasize that the incorporation and interaction of ID information can enhance the model's capability to distinguish similar trajectories in complex scenarios while facilitating better assignment decisions.
Incidentally, the introduction of the ID field also enables the trajectory augmentation mentioned in~\cref{sec:training}, further boosting the tracking performance, as shown in~\cref{tab:trajectory-augmentation}.

\vspace{2pt}
\noindent \textbf{Trajectory Augmentation.}
In~\cref{tab:trajectory-augmentation}, we explore the hyperparameters of the two different trajectory augmentation approaches mentioned in~\cref{sec:training}.
The performance significantly improves when $\lambda_\text{occ}$ is set to $0.5$. 
However, if too many tokens are discarded ($\lambda_\text{occ} = 1.0$), it can undermine the results due to excessive difficulty. 
Set $\lambda_\text{occ}$ to $0.5$, when progressively increasing the $\lambda_\text{sw}$ from $0.2$ to $0.8$, our method achieves the best performance while $\lambda_\text{sw}$ is set to $0.5$. 
Therefore, we use $\lambda_\text{occ} = \lambda_\text{sw} = 0.5 $ to conduct experiments as the final results in~\cref{sec:sota}.
It should be noted that using different augmentation hyperparameters on different datasets can yield better performance. However, to avoid over-focusing on engineering tricks, we use this unified setting across all datasets.

\begin{table}[t]
  \centering
  \setlength{\tabcolsep}{5pt}{
    \begin{tabular}{cc|ccccc}
    \toprule[2pt]
    $\lambda_{\textit{occ}}$ & $\lambda_{\textit{sw}}$ & HOTA & DetA & AssA & MOTA & IDF1 \\ 
    \midrule[1pt]
      0.0 & 0.0 & 59.5 & 75.3 & 47.2 & 85.6 & 61.1 \\
      0.5 & 0.0 & 60.7 & 75.0 & 49.4 & 85.2 & 62.7 \\
      1.0 & 0.0 & 58.6 & \bf 75.5 & 45.7 & \bf 85.7 & 59.8 \\
    \midrule[1pt]
      0.5 & 0.2 & 61.6 & 75.4 & 50.5 & 85.4 & 64.1 \\
      \rowcolor{gray!20} 0.5 & 0.5 & \bf 62.2 & 75.3 & \bf 51.5 & 85.2 & \bf 64.8 \\
      0.5 & 0.8 & 59.8 & 75.0 & 47.9 & 82.7 & 61.8 \\
    \bottomrule[2pt]
    \end{tabular}
  }
  \caption{Exploration of the hyperparameters for the trajectory augmentation techniques mentioned in~\cref{sec:training}. The \colorbox{gray!20}{gray} background is the choice for our final experiment.}
  \label{tab:trajectory-augmentation}
\end{table}
\section{Limitations and Discussions}
\label{sec:limitations-and-discusssions}

Although we achieved new state-of-the-art results across numerous datasets, our method still has considerable room for improvement, and several noteworthy limitations remain. 
As discussed in~\cref{sec:motip-architecture}, our approach is simple and intuitive, adhering to the philosophy \textit{less is more}. 
Therefore, our primary objective is to verify the feasibility of treating MOT as an in-context ID prediction process, rather than delving into highly customized model designs. 
This leaves ample room for future research to explore enhancements and customizations, such as tailored ID Decoder layers, additional tracking cues (\eg motion, depth, \etc), and more sophisticated trajectory modeling techniques. 
Another limitation is that the capacity $K$ of the ID dictionary may not be enough in crowded scenarios. We have shown that in most cases, the token utilization rate is below $40\%$. If necessary, $K$ can be adjusted upwards for extreme scenarios. Just like DETRs set the number of detect queries to $300$ by default, our setting is also for general scenarios.

\section{Conclusion}
\label{sec:conclusion}

We have introduced treating multiple object tracking as an in-context ID prediction task, which simplifies both the training and tracking processes. 
Based on this, we proposed MOTIP, a simple yet effective baseline design. Surprisingly, our method surpassed the state-of-the-art on all benchmarks.
This demonstrates the tremendous potential of our pipeline and method, suggesting it can serve as a viable inspiration for future research.

\section*{Acknowledgements}

This work is supported by the National Key R$\&$D Program of China (No. 2022ZD0160900), Jiangsu Frontier Technology R$\&$D Program (No. BF2024076), the Collaborative Innovation Center of Novel Software Technology and Industrialization, and Nanjing University-China Mobile Communications Group Co., Ltd. Joint Institute.
Besides, Ruopeng Gao would like to thank Yunzhe Lv for the kind discussion and Muyan Yang for the social support.

\appendix

\section{Overview}
\label{sec:supp-overview}

In the supplementary material, we primarily:

\begin{enumerate}
    \item State more experimental details, in~\cref{sec:supp-exp-details}.
    \item Discuss concerns regarding the introduction of static images for joint training, in~\cref{sec:supp-rethinking}.
    \item Provide additional experimental and visualization results, in~\cref{sec:supp-more-results}.
\end{enumerate}

\section{Experimental Details}
\label{sec:supp-exp-details}

Due to space constraints in the main text, we could not provide a comprehensive account of all experimental details. 
In this section, we will describe the specific details related to the training (\cref{sec:supp-training}), inference (\cref{sec:supp-inference}), and ablation experiments (\cref{sec:supp-ablation}).

\subsection{Training}
\label{sec:supp-training}

\noindent \textbf{Settings.}
In each training iteration, we need to sample $T+1$ frames, as mentioned in \cref{sec:implementation-details}.
Similar to previous works~\cite{MOTR, MeMOTR, CO-MOT} that employ multi-frame training, we adopt random sampling intervals to enhance the diversity of training data.
However, continuously increasing the sampling interval may make training samples excessively challenging. 
This could cause a discrepancy between training data and the inference video sequences, ultimately adversely affecting the model's performance. 
In our experiments, we set the random sampling interval to range from $1$ to $4$ by default. 

For the final training strategies, we have chosen the following approaches: 
On DanceTrack~\cite{DanceTrack}, we train MOTIP for $10$ epochs on the train set and drop the learning rate by a factor of $10$ at the $5$-th and $9$-th epoch. 
On SportsMOT~\cite{SportsMOT}, we train our model for $13$ epochs on the train set and drop the learning rate by a factor of $10$ at the $8$-th and $12$-th epoch. 
On BFT~\cite{NetTrack}, we train the model for $22$ epochs while drop the learning rate at the $16$-th and $20$-th epoch. 
To expedite the convergence, we use COCO~\cite{COCO} pre-trained weights and perform detection pre-training on the corresponding datasets. This serves as the initialization for the DETR part of MOTIP. 
Our typical hardware setup involves 8 NVIDIA RTX 4090 GPUs, with the batch size of each GPU set to $1$.

\begin{figure}[t]
    \centering
    \includegraphics[width=1.0\linewidth]{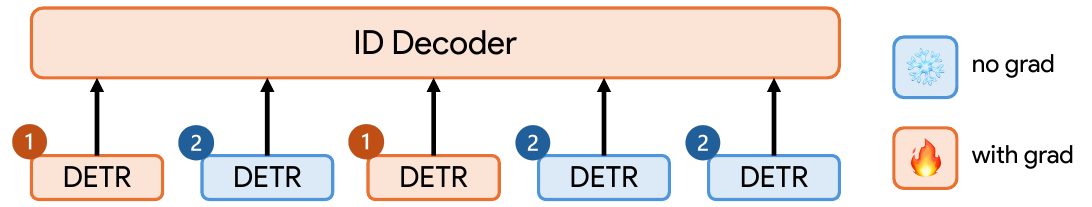}
    \caption{
    Illustration of the parallelized training of MOTIP, using a five-frame demo. 
    Since the detection process for each frame is independent, all DETRs in a sequence can perform forward simultaneously, which is GPU-friendly.
    In our implementation, we divide all DETRs into two forward passes (as shown in numbers $1$ and $2$) since we only backpropagate gradients for a subset of them, as described in \cref{sec:implementation-details} 
    }
\label{fig:supp-parallelized-training}
\end{figure}

\begin{figure}[t]
    \centering
    \includegraphics[width=1.0\linewidth]{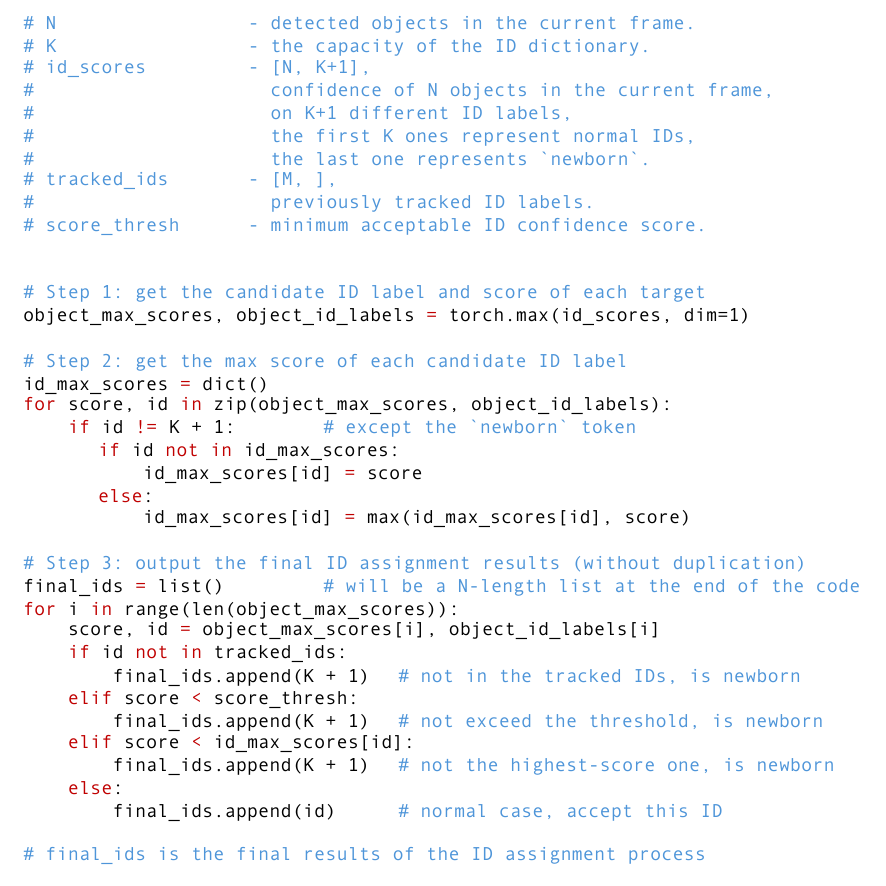}
    \caption{Python-like pseudocode for the core of our ID assignment process.}
\label{fig:supp-assignment}
\end{figure}

\begin{figure*}[t]
    \centering
    \includegraphics[width=1.0\linewidth]{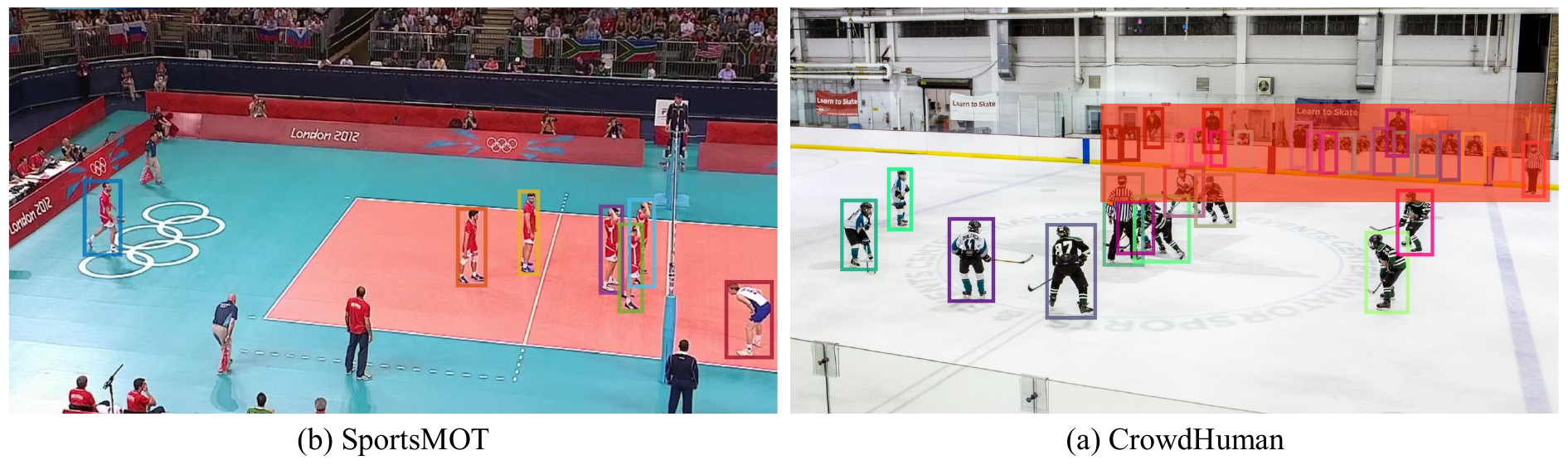}
    \caption{
    \textbf{Visualizing the different annotation standards between SportsMOT~\cite{SportsMOT} and CrowdHuman~\cite{CrowdHuman}.}
    (a) In SportsMOT, only athletes are annotated, excluding referees and spectators, or any other people. 
    (b) Since CrowdHuman aims to detect all humans, it additionally includes annotations for crowds outside the sports venues, as shown by the area covered in the \colorbox{red!20}{red mask}.
    }
\label{fig:supp-inconsistent-annotations}
\end{figure*}

\noindent \textbf{Parallelization.}
Recent tracking-by-query methods~\cite{MOTR, MeMOTR, CO-MOT}, which also use DETRs as their frameworks, process multi-frame video sequences in a manner similar to RNNs during training.
For instance, when processing a five-frame video clip, the model needs to perform five sequential forward passes of the DETR component, one frame at a time. 
Since the DETR architecture accounts for the majority of computational cost, this processing approach fails to leverage the parallel processing capabilities of the GPU.
In contrast, our MOTIP decouples detection and association components, allowing it to detect targets in all frames at once during training, as illustrated in~\cref{fig:supp-parallelized-training}.
Meanwhile, since our ID Decoder is a transformer decoder structure, it can also achieve parallelism by leveraging the attention masks~\cite{Attention}.
Therefore, our method can attain high parallelism on the GPU during training, improving GPU utilization and enabling efficient training.

\subsection{Inference}
\label{sec:supp-inference}

As discussed in \cref{sec:inference}, during inference, we utilize a straightforward ID assignment strategy. 
We apply an approach similar to classification tasks, selecting the prediction with the highest confidence score for each object as the final accepted result. 
This simplifies our inference process, eliminating the need for more intricate allocation strategies.
Although this approach seems feasible, most tracking evaluation approaches~\cite{MOTA, IDF1, HOTA} cannot handle duplicate IDs within the same frame. 
Therefore, we need to introduce an additional rule to handle this situation: when duplicate IDs appear in the final results, we select the one with the highest confidence and label the others as newborn objects. 
This simple patch completely avoids the occurrence of duplicate ID labels within the same frame.
The pseudocode for the aforementioned assignment process is provided in~\cref{fig:supp-assignment}.

\begin{figure*}[t]
    \centering
    \includegraphics[width=1.0\linewidth]{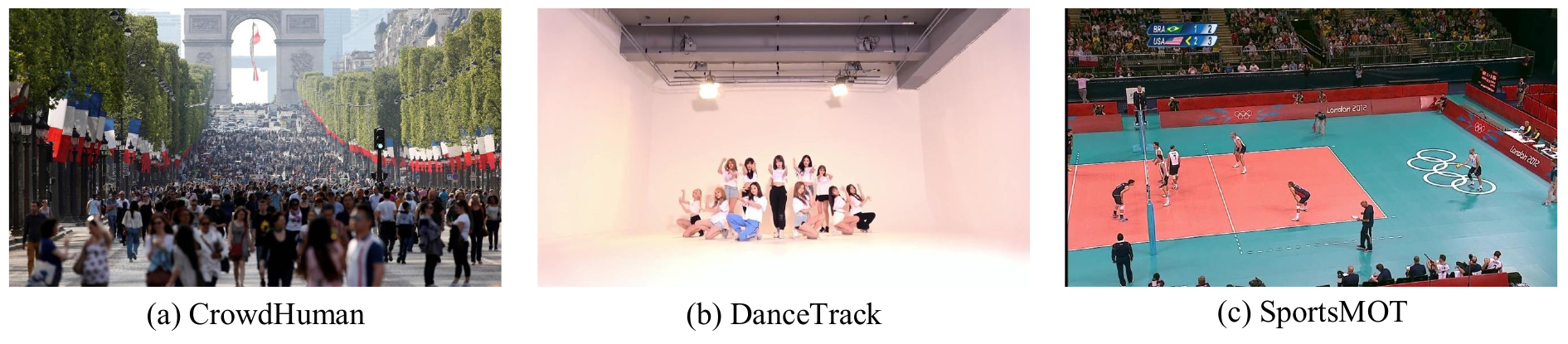}
    \caption{\textbf{Illustrate the inconsistent scenario characteristics from different datasets.} (a) humans in high-density scenarios from CrowdHuman~\cite{CrowdHuman}. (b) DanceTrack~\cite{DanceTrack} aims to track indoor dancers. (c) SportsMOT~\cite{SportsMOT} is chiefly concerned with the tracking of sports events. }
\label{fig:supp-inconsistent-scenarios}
\end{figure*}

\begin{figure*}[tb]
  \centering
  \begin{subfigure}{1.0\linewidth}
    \includegraphics[width=1.0\linewidth]{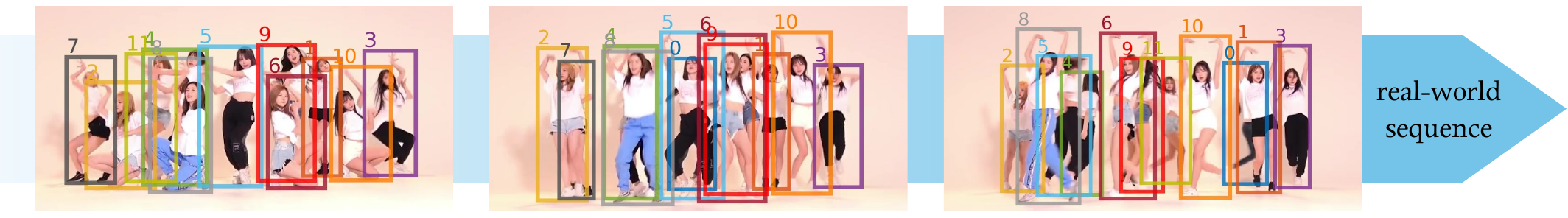}
    \caption{A real-world video sequence, which is directly sampled from DanceTrack~\cite{DanceTrack}.}
    \label{fig:supp-dancetrack-sequence}
  \end{subfigure}
  \\
  \begin{subfigure}{1.0\linewidth}
    \includegraphics[width=1.0\linewidth]{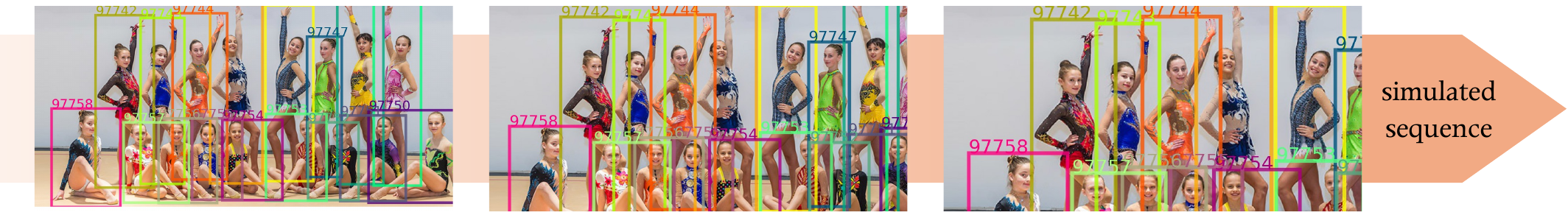}
    \caption{A simulated video sequence is generated by sampling regions through a random shift technique~\cite{MOTR} from a static image (in CrowdHuman~\cite{CrowdHuman}).}
    \label{fig:supp-crowdhuman-sequence}
  \end{subfigure}
  \caption{
  Illustrating two distinct approaches of video sequence acquisition: \textit{real-world vs. simulated} sequences. 
  The latter is tantamount to transform the objects by mere translational and scaling transformations, which intuitively seems overly simplistic for a tracking model. 
  }
  \label{fig:supp-sequence}
\end{figure*}

\subsection{Ablation Studies}
\label{sec:supp-ablation}

\noindent \textbf{Default Settings. }
As stated in \cref{sec:ablation}, we train the models on the train set of DanceTrack~\cite{DanceTrack} and subsequently evaluate on its official validation set to conduct our ablation experiments. 
Unless otherwise stated, all experiments are implemented without using trajectory augmentation techniques to ensure a fair comparison, \ie, $\lambda_\text{occ} = \lambda_\text{sw} = 0.0$.
To reduce the computational cost, we shortened the sampling length of the video sequence to $T = 19$. 
As for the inference, we leverage $\lambda_{\textit{det}}=\lambda_{\textit{new}}=0.5$, while $\lambda_{\textit{id}}=0.1$ for simplicity.
Other unspecified details are consistent with the default setups, as referenced in \cref{sec:implementation-details}, ~\cref{sec:supp-training} and~\cref{sec:supp-inference}.

\vspace{2pt}
\noindent \textbf{\textit{Re-ID} Pipeline.}
As shown in Tab.~5, we construct a \textit{re-id} pipeline to compare with our in-context ID prediction approach. 
During the training process, we refer to FairMOT~\cite{FairMOT}, a well-known joint detection and embedding method. 
Interestingly, it also utilized the cross-entropy function for supervision. This similarity undoubtedly provides a suitable competitor for our method. 
In practice, we treat each trajectory as a class and assign it a unique ID label that remained consistent throughout the entire training process. 
After obtaining the output embeddings from the model, we employ a linear projection as the classification head and use the corresponding ID labels to supervise the classification results of the targets. 
Although both approaches use classification supervision, our procedures for defining ID labels differ significantly, as discussed in \cref{sec:in-context-ID-prediction}.
During inference, the \textit{re-id} pipeline applies cosine similarity to calculate the cost matrix for ID assignment, which is a widely adopted strategy in ReID-based methods~\cite{FairMOT, Hybrid-SORT, Deep-OC-SORT}.

\vspace{2pt}
\noindent \textbf{\textit{Contra} Pipeline.}
Inspired by contrastive learning methods like CLIP~\cite{CLIP}, recent work~\cite{ConstrastiveDETR-MOT} has employed contrastive learning to supervise the object embeddings of different trajectories, aiming to learn distinguishable features. 
Therefore, we employ the infoNCE loss~\cite{CLIP, ConstrastiveDETR-MOT} to supervise the model as a comparative method, which is denoted as \textit{contra} pipeline in Tab.~5. 
In experiments, we also tune some hyperparameters to improve the performance (about $3.0$ HOTA) for a thorough comparison.
During inference, we also use the cosine similarity matrix.
In the above two comparative pipelines, we observe that the tracking performance is similar when the similarity threshold is below $0.5$.
Therefore, we retain the similarity threshold at $0.1$ in our ablation experiments for simplicity.

\vspace{2pt}
\noindent \textbf{One-Stage and Two-Stage Training}
In Tab.~5, we establish two different training strategies: \textit{one-stage} and \textit{two-stage}. 
The former uses a combined loss function to supervise the entire network, like Eq.~(3).
In contrast, two-stage training divides the model into two sequentially trained parts.
First, the DETR component is trained using detection supervision. Then, the trained weights are frozen, and the object association part is trained separately.
This two-stage training ensures the consistency of the object embeddings produced by DETR, thereby providing a fair testbed for the three different pipelines.

\section{Rethinking Joint Training with Images}
\label{sec:supp-rethinking}

Some recent work~\cite{MOTRv2, CO-MOT, MOTRv3, SportsMOT} has opted to use additional image detection datasets, such as CrowdHuman~\cite{CrowdHuman}, for joint training. 
For methods~\cite{MOTRv2, CO-MOT, MOTRv3} requiring multi-frame training, they commonly use random shifting to simulate short video clips from single images to meet the training requirements. 
While this kind of joint training can indeed improve tracking performance, we argue it hinders the sustained advancement of MOT methods, particularly for end-to-end models that focus on temporal information. 
The main impacts are concentrated in three areas: inconsistent scene characteristics (\cref{sec:supp-inconsistent-scenarios}), inconsistent annotation standards (\cref{sec:supp-inconsistent-annotations}), and overly simplistic video simulations (\cref{sec:supp-simplistic-simulations}).

\subsection{Inconsistent Scenario Characteristics}
\label{sec:supp-inconsistent-scenarios}

As discussed in Sec. 3.1 of CrowdHuman~\cite{CrowdHuman}, this dataset aims to be diverse for real-world scenarios.
To this end, various different keywords were used to collect data from Google Image search.
In contrast, existing MOT benchmarks~\cite{DanceTrack, SportsMOT, MOT16} predominantly focus on specific scenarios.
For instance, SportsMOT~\cite{SportsMOT} primarily collects high-quality videos from professional sports events, while DanceTrack~\cite{DanceTrack} crawls network videos, including mostly group dancing.
Consequently, in CrowdHuman, some scenes may never appear in specific MOT datasets. 
As illustrated in~\cref{fig:supp-inconsistent-scenarios}, some crowded scenes of CrowdHuman are virtually absent in both DanceTrack and SportsMOT.
Additionally, the CrowdHuman dataset also encompasses some scenes under atypical low-light conditions and wide-angle lens perspectives, which significantly deviate from the distribution of the target datasets, like DanceTrack and SportsMOT. 

Although the inconsistencies across these scenarios did not adversely impact the performance on target benchmarks yet, there are still some concerns based on common consensus in deep learning training. 
Using out-of-domain data is inherently a double-edged sword.
While it can boost performance, it may also cause the model astray from its intended application scenarios.
This typically calls for a careful adjustment of the training data ratio during training to maintain this delicate balance.
We argue this would lead researchers to spend an excessive amount of unnecessary effort on tuning the hyperparameters required for training. 
While additional detection datasets are still necessary for some traditional, smaller datasets (like MOT15~\cite{MOT15}, MOT17~\cite{MOT16}), many recently proposed benchmarks already contain sufficient training data to fully unlock the potential of most models. Introducing extra image data, in this case, would be redundant.

\subsection{Inconsistent Annotation Standards}
\label{sec:supp-inconsistent-annotations}

Across different datasets, even if the category labels seem the same, the focal points may vary.
A notable example is the distinction between the CrowdHuman~\cite{CrowdHuman} and SportsMOT~\cite{SportsMOT} datasets.
CrowdHuman aims to detect every visible person in the images, whereas SportsMOT focuses solely on the athletes in the videos. 
This results in differences in the annotation protocols between these two datasets. 
As illustrated in~\cref{fig:supp-inconsistent-annotations}, compared to SportsMOT, CrowdHuman includes additional annotations for spectators and referees in sports scenes. 
Using these two datasets for joint training can confuse the model because of the different annotation standards. Specifically, it becomes unclear if people who are not athletes should be considered as positive detections, thereby impairing the final performance. 
Some more sophisticated engineering designs have been used to address this issue. For example, MixSort~\cite{SportsMOT} employs different combinations of training data at multiple stages and finally performs fine-tuning exclusively on SportsMOT. 
However, the optimal joint training strategy can vary for different models while dealing with this issue. 
We argue that customizing multi-stage joint training strategies would divert researchers' efforts toward engineering tricks rather than general tracking solutions.
Therefore, we believe that when the amount of training data is sufficient to validate the effectiveness of the method, there is no need to introduce additional training data. This helps avoid the complexity of adjusting training strategies.

\subsection{Simplistic Video Simulations}
\label{sec:supp-simplistic-simulations}

Recent research~\cite{MOTR, MeMOTR, SambaMOTR} has demonstrated that multi-frame training is highly beneficial for developing a more robust tracking model.
When incorporating detection datasets like CrowdHuman~\cite{CrowdHuman} for joint training, random shifting is employed to sample different regions of the same image to generate video clips. 
For each target, this is equivalent to continuously applying a translation and scaling operation at a constant ratio, as shown in~\cref{fig:supp-crowdhuman-sequence}.
The resulting sequence will have very small differences between frames, lacking features such as deformation, occlusion, and changes in relative position that are present in real-world sequences (as shown in~\cref{fig:supp-dancetrack-sequence}). 
This phenomenon becomes more pronounced as the sampling length increases. Because the usable area of the static image remains constant, the shifting scale of each step must be reduced to ensure validity, making adjacent frames even more similar. 
Recent work~\cite{MeMOTR, MOTR, SambaMOTR, QuoVadis} has increasingly focused on the application of temporal information in tracking and has benefited from long-term sequence training.
However, the overly simplistic method of video simulation can contaminate the distribution of training data, thereby severely impairing model performance.
This might impede researchers from delving deeper into the exploration of temporal information. 
While more complex and diverse video simulation approaches can help alleviate this issue, this goes beyond the scope of general MOT methods. For more details, you can refer to some related studies~\cite{MASA, TrackDiffusion, OVTrack}.

\subsection{Discussions}
\label{sec:supp-rethinking-discussions}

Our MOTIP is an end-to-end learnable approach, which encounters additional challenges during joint training, as discussed in~\cref{sec:supp-inconsistent-scenarios} and~\cref{sec:supp-inconsistent-annotations}. Additionally, we utilize long-sequence training to handle temporal information, which means that video data generated through image simulation does not sufficiently benefit our model, as discussed in~\cref{sec:supp-simplistic-simulations}.
For example, during joint training of SportsMOT~\cite{SportsMOT} and CrowdHuman~\cite{CrowdHuman}, single-stage training can cause issues for the detector due to annotation inconsistencies. On the other hand, multi-stage training struggles to address the problem of model forgetting. 
This challenge is not unique to our model.
As the computer vision community evolves, more end-to-end and long-term modeling approaches will become available for multi-object tracking. 
Complex joint training might shift the focus towards engineering implementations rather than the research of more generalized tracking methods.
Therefore, we suggest that when the dataset is sufficiently large, it may be preferable to avoid introducing extra data for training. This approach allows for a more focused effort on addressing the various challenges in multiple object tracking.

\section{More Results}
\label{sec:supp-more-results}

\begin{table}[t]
  \centering
  \setlength{\tabcolsep}{3.5pt}{
    \begin{tabular}{l|ccccc}
      \toprule[2pt]
      Methods & HOTA & DetA & AssA & MOTA & IDF1 \\
      \midrule[1pt]
      \textit{heuristic:} \\
          CenterTrack~\cite{CenterTrack} & 52.2 & 53.8 & 51.0 & 67.8 & 64.7 \\
          QDTrack~\cite{QDTrack} & 53.9 & 55.6 & 52.7 & 68.7 & 66.3 \\
          GTR~\cite{GTR} & 59.1 & 61.6 & 57.0 & 75.3 & 71.5 \\
          FairMOT~\cite{FairMOT} & 59.3 & 60.9 & 58.0 & 73.7 & 72.3 \\
          DeepSORT~\cite{Deep-SORT} & 61.2 & 63.1 & 59.7 & 78.0 & 74.5 \\
          SORT~\cite{SORT} & 63.0 & 64.2 & 62.2 & 80.1 & 78.2 \\
          ByteTrack~\cite{ByteTrack} & 63.1 & 64.5 & 62.0 & 80.3 & 77.3 \\
          OC-SORT~\cite{OC-SORT} & 63.2 & 63.2 & 63.4 & 78.0 & 77.5 \\
          C-BIoU~\cite{C-BIoU} & 64.1 & 64.8 & 63.7 & 81.1 & 79.7 \\
          MotionTrack~\cite{MotionTrack} & 65.1 & 65.4 & 65.1 & 81.1 & 80.1 \\
      \midrule[1pt]
      \textit{end-to-end:} \\
          TrackFormer~\cite{TrackFormer} & / & / & / & 74.1 & 68.0 \\
          MeMOT~\cite{MeMOT} & 56.9 & / & 55.2 & 72.5 & 69.0 \\
          MOTR~\cite{MOTR} & 57.2 & 58.9 & 55.8 & 71.9 & 68.4 \\
          \textcolor{gray}{MOTRv2$^\dag$~\cite{MOTRv2}} & \textcolor{gray}{57.6} & \textcolor{gray}{58.1} & \textcolor{gray}{57.5} & \textcolor{gray}{70.1} & \textcolor{gray}{70.3} \\
          \textcolor{gray}{MeMOTR~\cite{MeMOTR}} & \textcolor{gray}{58.8} & \textcolor{gray}{59.6} & \textcolor{gray}{58.4} & \textcolor{gray}{72.8} & \textcolor{gray}{71.5} \\
          MOTIP~(ours) & \bf 59.3 & \bf 62.0 & \bf 57.0 & \bf 75.3 & \bf 71.3 \\
          \textcolor{gray}{CO-MOT~\cite{CO-MOT}} & \textcolor{gray}{60.1} & \textcolor{gray}{59.5} & \textcolor{gray}{60.6} & \textcolor{gray}{72.6} & \textcolor{gray}{72.7} \\
          \textcolor{gray}{MOTRv3~\cite{MOTRv3}} & \textcolor{gray}{60.2} & \textcolor{gray}{62.0} & \textcolor{gray}{56.9} & \textcolor{gray}{75.5} & \textcolor{gray}{71.2} \\
      \bottomrule[2pt]
    \end{tabular}
  }
  \caption{
  Performance comparison with state-of-the-art methods on MOT17~\cite{MOT16}. The best performance among the end-to-end methods is marked in \textbf{bold}. 
  The results shown in \textcolor{gray}{gray font} indicate unfair comparisons due to network structure, as we detailed in~\cref{sec:supp-mot17}.
  MOTRv2$^\dag$ refers to the results of MOTRv2~\cite{MOTRv2} after removing additional heuristic post-processing algorithms, as derived from~\cite{MOTRv3}. 
  }
  \label{tab:supp-mot17}
\end{table}

\subsection{MOT17}
\label{sec:supp-mot17}

Although MOT17~\cite{MOT16} is widely recognized as an important pedestrian tracking dataset, its limited amount of training data has been noted by many works~\cite{MeMOTR, SambaMOTR} to be inadequate for training modern models, especially end-to-end approaches. 
Since it only contains $7$ video sequences for training, current methods~\cite{ByteTrack, MOTR} always incorporate extra detection datasets~\cite{CrowdHuman, MOT15} for joint training to ensure data diversity. 
Nonetheless, some studies~\cite{MeMOTR} have shown that the lack of diversity makes models prone to overfitting on training data, resulting in insufficient generalization capabilities. 
Under the same settings, end-to-end methods face more severe problems compared to heuristic algorithms, because additional datasets are insufficient for models to learn optimal tracking strategies, as we discussed in~\cref{sec:supp-simplistic-simulations}.
We argue these compromises and issues might divert research from fundamental tracking solutions, causing an overemphasis on engineering details.
For this reason, we chose some more modern and diverse datasets in our main text, such as DanceTrack~\cite{DanceTrack}, SportsMOT~\cite{SportsMOT}, and BFT~\cite{NetTrack}, to ensure the model is well-trained. 

Nevertheless, we still present the state-of-the-art comparison on MOT17~\cite{MOT16} in~\cref{tab:supp-mot17}.
To handle crowded scenes, we modify several hyperparameters, such as setting the capacity of the ID dictionary ($K$) to $200$.
Compared to MOTR~\cite{MOTR}, which also uses the standard Deformable DETR framework, our MOTIP shows a significant performance improvement ($59.3$ HOTA \textit{vs.} $57.2$ HOTA).
It should be noted that some of the methods (in \textcolor{gray}{gray font}) in~\cref{tab:supp-mot17} are not a fair comparison with ours: MeMOTR~\cite{MeMOTR} uses DAB-Deformable DETR~\cite{DAB-DETR} as the framework, while CO-MOT~\cite{CO-MOT} customizes the reference points in Deformable DETR. MOTRv2$^\dag$ employs an additional YOLOX detector~\cite{YOLOX} as the proposal generator. MOTRv3 utilizes a more powerful backbone, ConvNeXT-Base~\cite{ConvNeXT}.
Nevertheless, compared to these latest work, our method still demonstrates competitive performance.
However, there is still a significant gap between our method and the state-of-the-art heuristic algorithms. 
On the one hand, heuristic algorithms have been continuously customized and developed over the past decade for these linear motion scenarios. In contrast, end-to-end approaches lack this human-crafted prior knowledge and still require time to mature. 
On the other hand, as previously discussed, overly homogeneous training data is detrimental to learnable methods.
Therefore, we look forward to diverse and large-scale pedestrian tracking datasets to better explore and evaluate end-to-end general tracking methods.

\subsection{Inference Speed}
\label{sec:supp-speed}

\begin{table}[t]
  \centering
    \setlength{\tabcolsep}{6pt}{
    \begin{tabular}{c|c|c}
    \toprule[2pt]
    Methods & FP32 & FP16 \\
    \midrule[1pt]
    MOTR~\cite{MOTR}   & 13.1 FPS   & / \\
    MOTIP (ours)   & 12.7 FPS   &  22.8 FPS \\
    \bottomrule[2pt]
    \end{tabular}
    }
  \caption{
  Comparison of inference speed.
  The experiments are conducted on a single NVIDIA RTX A5000 GPU. Using FP16 precision, MOTIP can achieve near real-time performance.
  }
  \label{tab:supp-inference-speed}
\end{table}

Based on the analysis of our network structure, although we introduce an ID Decoder structure, its computational cost during inference is negligible compared to that required by Deformable DETR~\cite{DeformableDETR}. 
In~\cref{tab:supp-inference-speed}, we compare our method with another query-based approach that also uses Deformabe DETR. The results show that MOTIP and MOTR~\cite{MOTR} have similar inference speeds, supporting our perspective. 
Additionally, we are surprised to find that at FP16 precision, MOTIP can achieve nearly real-time inference speed, indicating its feasibility for practical applications.
To address real-time considerations in the future, Deformable DETR could be replaced with some recent real-time DETR frameworks~\cite{RT-DETR, LW-DETR}.

\subsection{Visualization of ID Decoder}
\label{sec:supp-visualization-ID-decoder}

\begin{figure*}[t]
    \centering
    \includegraphics[width=1.0\linewidth]{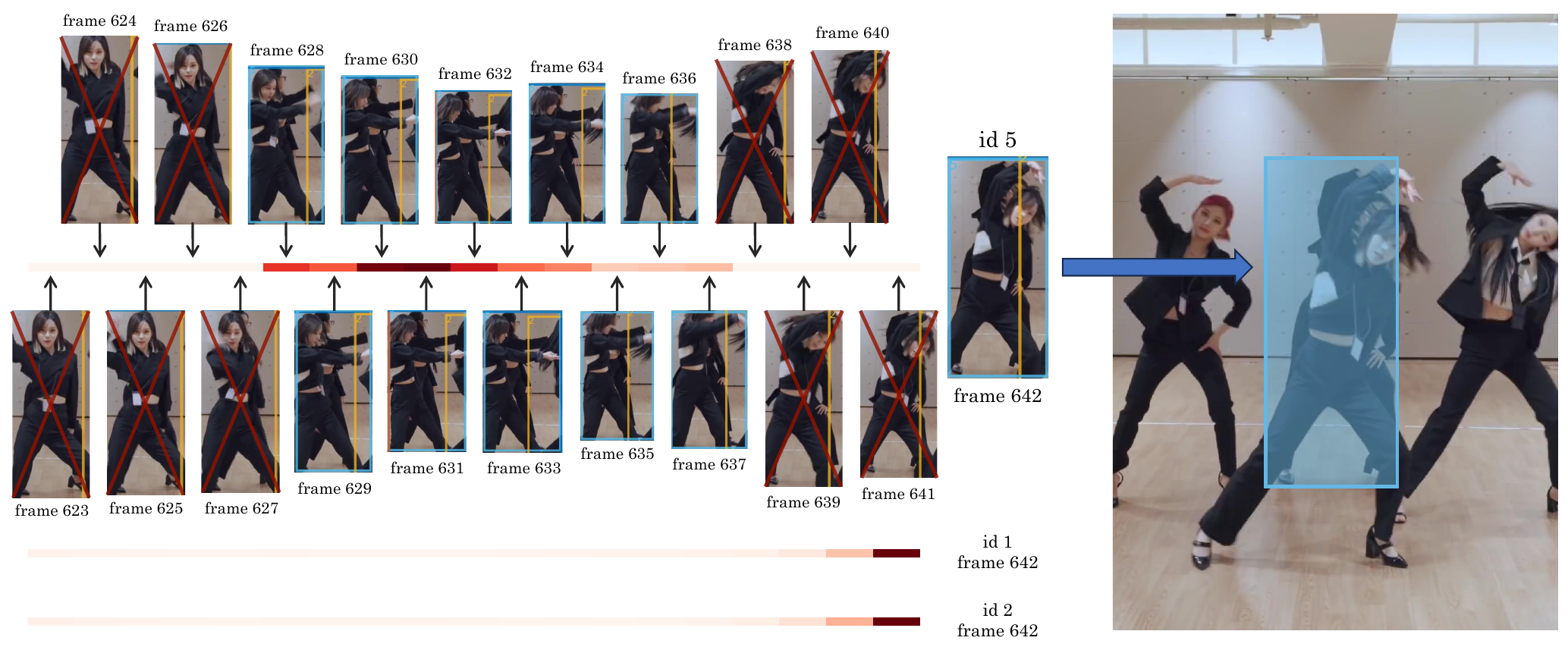}
    \caption{
    \textbf{Visualization of the cross-attention scores in the ID Decoder.}
    We show the response intensity between a target and its corresponding historical tracklets, with darker shades indicating stronger responses. 
    Object $5$ is occluded from frame $638$ to $641$ and reappears in frame $642$. The other two objects, $1$ and $2$, remain visible during these $20$ frames. 
    The targets marked with a \textcolor{red}{red cross} indicate that they are not visible in the current frame.
    For a more comprehensive example, please refer to~\cref{fig:supp-decoder-attn-full}.
    }
\label{fig:supp-decoder-attn}
\end{figure*}

\begin{figure*}[t]
    \centering
    \includegraphics[width=0.8\linewidth]{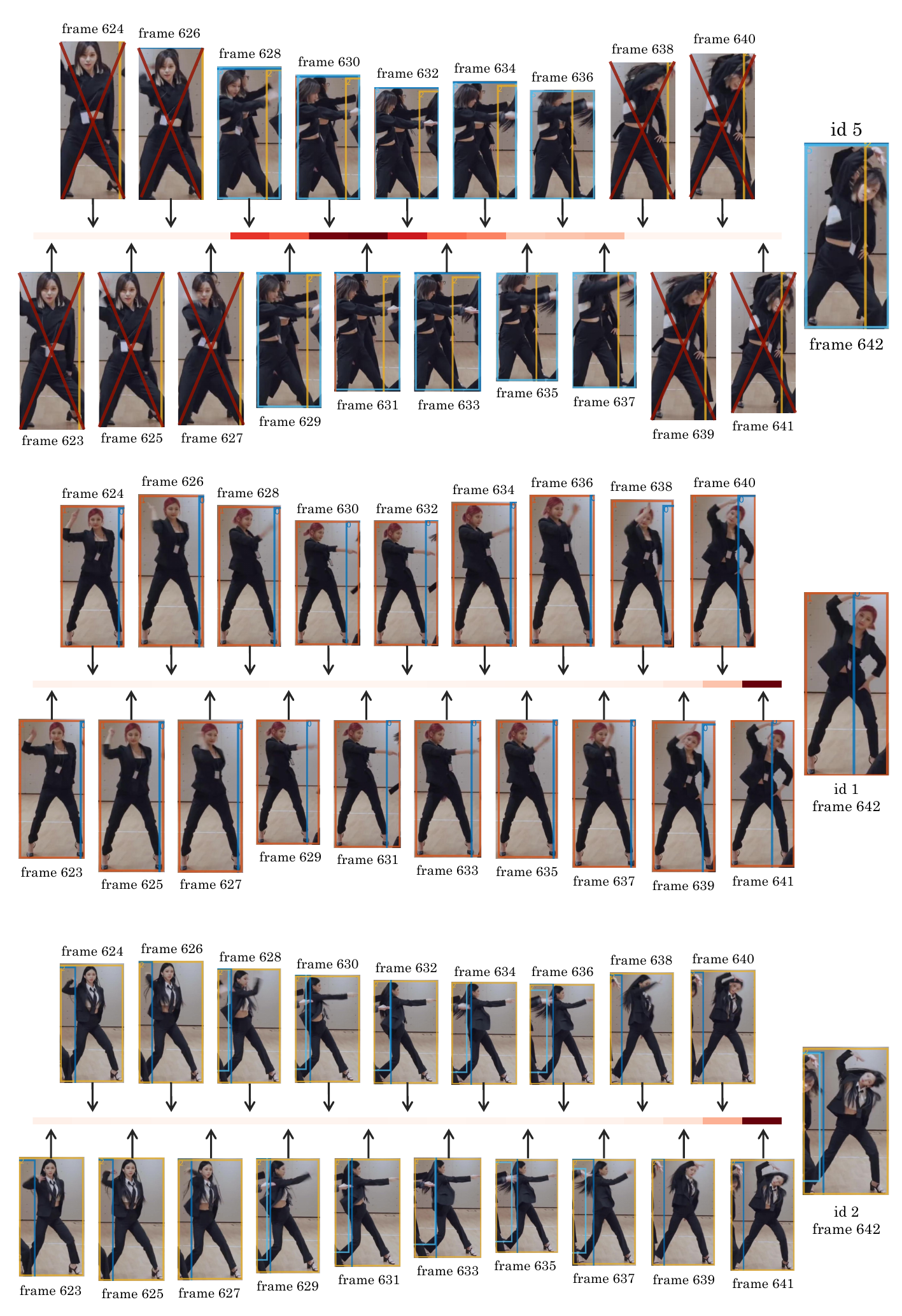}
    \caption{
    \textbf{A more comprehensive illustration of the example in~\cref{fig:supp-decoder-attn}.}
    Unlike object $5$, targets $1$ and $2$ are clearly visible in all frames, and therefore, naturally select the closest results as reliable features. 
    }
\label{fig:supp-decoder-attn-full}
\end{figure*}

As discussed in \cref{sec:ablation}, we believe MOTIP is more flexible and intelligent in handling historical trajectory information compared to heuristic-based Re-ID methods. 
In this section, we use some visualizations to elucidate this explanation. 
In~\cref{fig:supp-decoder-attn}, we present a case where target $5$ is not visible from frames $638$ to $641$ and reappears in frame $642$. 
When a target disappears, it is always accompanied by severe occlusion issues. 
For example, as shown in~\cref{fig:supp-decoder-attn}, although the detector can correctly identify target $5$ in frame $636$, she is almost completely occluded by the dancer standing in front. 
This severe occlusion can render the target features unreliable. 
In traditional heuristic algorithms, this issue cannot be dynamically identified and addressed because the matching rules are manually fixed. 
However, our MOTIP can make dynamically optimal choices in such situations based on its cross-attention structure. 
As shown in~\cref{fig:supp-decoder-attn}, object $5$ selects the targets in frames $630$ and $631$ as more reliable features, avoiding the pitfalls of unreliable ones (like in frame $640$ and $641$).
In contrast, targets $1$ and $2$, which are not occluded throughout, will select the features closest to the current frame as they are the most similar. 
For a more detailed illustration, please refer to~\cref{fig:supp-decoder-attn-full}.
We believe that compared to manually crafted rules based on experience, this flexible dynamic decision-making learned directly from data can help the model make accurate choices in challenging scenarios.

\subsection{Number of ID Decoder Layers}
\label{sec:supp-decoder-layers}

\begin{table}[t]
  \centering
  \setlength{\tabcolsep}{4pt}{
    \begin{tabular}{c|ccccc}
    \toprule[2pt]
    \textit{Dec Layers} & HOTA & DetA & AssA & MOTA & IDF1 \\ 
    \midrule[1pt]
     1 & 54.3 & 75.5 & 39.3 & 84.6 & 51.9 \\
     3 & 57.6 & \bf 75.8 & 44.0 & \bf 85.8 & 58.3 \\
     \rowcolor{gray!20} 6 & 59.5 & 75.3 & 47.2 & 85.6 & 61.1 \\
     9 & \bf 60.5 & 75.3 & \bf 48.9 & 85.7 & \bf 62.4 \\
     12 & 60.3 & 75.0 & 48.7 & 85.1 & 61.7 \\
    \bottomrule[2pt]
    \end{tabular}
  }
  \caption{Ablation experiments on the number of layers in the proposed ID Decoder. The \colorbox{gray!20}{gray} background is the choice for our final experiment.}
  \label{tab:supp-decoder-layers}
\end{table}

As a key component, we investigate the impact of different numbers of layers in the ID Decoder on the final tracking performance in~\cref{tab:supp-decoder-layers}. 
Overall, as the number of layers increases, the final tracking performance improves gradually (from $54.3$ to $60.5$ HOTA). 
We believe this is because more decoding layers allow for elaborate modeling and further refinements of the ID allocations, enabling the model to handle more complicated situations.
However, empirical evidence suggests an excessive number of network layers may lead to difficulties in model convergence, thereby increasing the training burden.
At the same time, the improvements brought by increasing decoding layers exhibit diminishing marginal returns.
Based on the above considerations, as mentioned in \cref{sec:implementation-details}, we select a $6$-layer structure as our default configuration.

\subsection{Previous Results}
\label{sec:supp-previous-results}

This paper has an earlier version at \href{https://arxiv.org/abs/2403.16848v1}{arXiv:2403.16848v1}.
Although the model structure remains unchanged, we updated the codebase and some hyperparameters, resulting in improved tracking performance in \cref{sec:sota} compared to the earlier version.
We suggest that, for subsequent studies, comparing either of these two results based on their respective code frameworks is reasonable and acceptable.

{
    \small
    \bibliographystyle{ieeenat_fullname}
    \bibliography{main}
}


\end{document}